# SMARTPHONE-BASED METHOD FOR AUTOMATED SPEED ENFORCEMENT

**Keya Li**
Graduate Research Assistant
Department of Civil, Architectural and Environmental Engineering
The University of Texas at Austin
keya_li@utexas.edu

**Jahnavi Malagavalli**
Graduate Research Assistant
Department of Computer Science
The University of Texas at Austin
jahnavimalagavalli@utexas.edu

**Lamha Goel**
Graduate Research Assistant
Department of Computer Science
The University of Texas at Austin
lamhag@utexas.edu

**Tong Wang**
Graduate Research Assistant
Chandra Department of Electrical and Computer Engineering
The University of Texas at Austin
tong.wang@utexas.edu

**Kara M. Kockelman, Ph.D., P.E.** (corresponding author)
Dewitt Greer Centennial Professor in Engineering
Department of Civil, Architectural and Environmental Engineering
The University of Texas at Austin
kkockelm@mail.utexas.edu 512-471-0210

## ABSTRACT

Smartphone cameras and computer vision (CV) hold significant promise in assisting public agencies with enforcing traffic laws and enhancing road safety. This work designs and tests a smartphone-based method for automated speed estimation and vehicle identification (license plate, make/model, and color recognition) via an automated pipeline to assist enforcement agencies in reliably identifying speeders. The CV code accurately recognizes nearly half (46%) of the license plates' text on 1,800 images from a Brazil open-source dataset, called UFPR-ALPR. Code tests on daytime recordings from hand-held smartphone videos (n=73) and roadside cameras (n = 42) in Austin, Texas yield 60.8% accuracy for color detection (among all possible RGB color categories), 48.6% on vehicle make/manufacturer identification, and 16.89% on vehicle make and model identification. Prediction accuracy for speed estimation (within a 20% range), vehicle make (within the top 3 predictions), and license plate recognition (within the top 10 predictions) are 16.3%, 16.9%, and 29.7%, respectively. This paper also illuminates the legal, technological, and practical aspects of using smartphones for enforcement, including the potential use of recordings for enforcement purposes, emphasizing the need to transform the potential of smartphone-based CV technologies into practical tools for vital information on traffic violations.

## 1. INTRODUCTION

Nearly 1.2 million people die each year globally in road traffic crashes (WHO, 2023), and almost 40,900 were killed on U.S. roadways in 2023 (NCSA, 2023). Speeding is a major contributor to U.S. crash counts and severities, with 28.7% of fatalities related to speeding (in 2021, and 29.3% in 2020) (Stewart, 2023). Active and automated enforcement of speed limits, road design strategies (like speed humps and purposeful use of red lights), speed governors on vehicles, and built-in tracking devices (like electronic logging devices) can significantly lower speeds, crash counts, and injuries (Sadeghi et al. 2016; Distefano and Leonardi, 2019; Rakesh, 2024). Statistics indicate that safety countermeasures like fixed speed cameras can reduce 47% of fatal and injury crashes on urban principal arterials, and variable speed limits can reduce up to 51% of crashes on freeways (Shin et al., 2009; Al-Marafi et al., 2020). It is difficult and costly to enforce driving laws using police officers in real-time, and speeders may out-race police cars, sometimes resulting in serious crashes (Rivara and Mack, 2004). Enforcement agencies have limited resources to allocate staff to identify violators and issue tickets on-site. One of the solutions is intelligent speed adaptation (ISA) which becomes mandatory for all new vehicle models sold from July 2024 across Europe (European Commission, 2024). This regulation aims to reduce collisions by 30% and prevent 140,000 serious road injuries by 2038, a significant step toward a goal of zero road deaths by 2050. More and more communities are turning to automated enforcement techniques to prosecute traffic violations, but such applications remain very rare in the U.S.

Automation of fee collection is widely used in various transportation settings, including automated collection of road tolls (in the U.S., Singapore, London, Italy, and elsewhere), identification of illegally parked vehicles (in almost any developed-nation city setting) (Kashid and Pardeshi, 2014), identification of reported-as-stolen vehicles (in the U.K., U.S., China, and elsewhere) (Farr et al., 2020; Chang and Su, 2010), and enforcement of speed limits and red-light compliance (in nearly half of U.S. states, EU, China, and elsewhere) (Heiny et al., 2023; Gössel, 2015). Automated noise-limit enforcement was recently implemented in locations around Paris, London, Taiwan, New York, and elsewhere, relying on radar detection (the radar device, composed of four microphones, measured noise levels every tenth of a second and triangulated the source of the sound) plus cameras for license plate reading (Moynihan and Esteban, 2019; NYC.gov, 2022). Presently, Russia may lead the world in speed enforcement deployments, with 18,413 speed cameras installed (Statista Research Department, 2023). As of July 2024, 247 communities across the U.S. have implemented speed safety programs (Insurance Institute for Highway Safety, 2024). As shown in Figure 1, 19 U.S. states and the District of Columbia permit the use of speed cameras (Governors Highway Safety Associate, 2024). These rely on radar waves and automatic number plate recognition (ANPR) programs for speed inference and license plate reading, either at single-camera stations (most common) or between cameras set miles apart (UK Department for Transport, 2007), and can be effective in reducing injuries and alleviating regulatory burdens. As of December 2021, thanks to the NYC Automated Speed Enforcement Program, speeding at fixed camera locations had dropped, on average, 73 percent in 750 school zones on all weekdays between 6 AM and 10 PM (NYC, 2023). Seattle's Speed Safety Camera Program reports 18% fewer pedestrian and bicyclist injury crashes at 17 camera sites/segments and 5% fewer along 100 adjacent segments (Heiny et al., 2023).

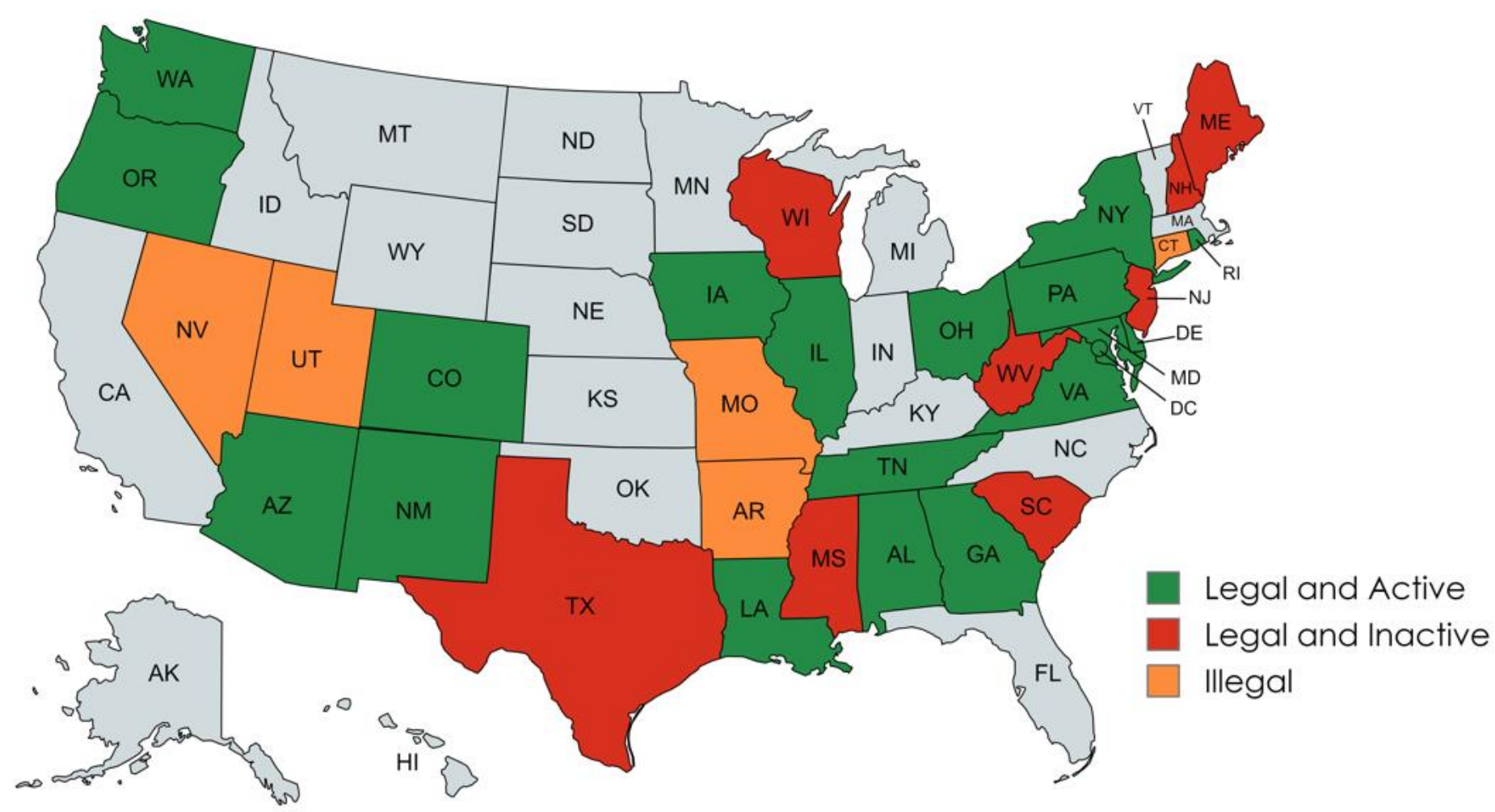

**Figure 1: Speed Cameras Use across the U.S. (as of July, 2024)**

Despite their road-safety and law-enforcement effectiveness, stationary cameras for reliable, automated traffic surveillance are expensive to purchase and maintain. According to New York City's Independent Budget Office (2016), speed camera costs averaged roughly $120,000 for hardware plus installation, and over $150,000 for 5 years of operation and maintenance. Given these costs, cities and states cannot afford to install stationary cameras along all road segments. Moreover, point cameras can cause drivers to slow down when being watched and then speed up downstream, a so-called 'kangaroo effect' (Chen et al., 2020) which reduces effectiveness. Instead, allowing private citizens to share videos of violations can assist in ensuring better driving at all times and in all settings. Citizens have been helping New York City officials enforce diesel-truck idling laws for several years; those submitting 3 minutes of video also receive 25 percent of any fine obtained from heavy-truck owners, which is close to $87.50 (Wilson, 2022).

This paper designs a smartphone-based method for speed enforcement, which includes automated/online speed inference and vehicle identification for automated reporting. This comprehensive framework allows for automatic detection of speeding vehicles and can output speeds and other information (such as license plate, color, make, and model) for delivery of information to public enforcement agencies. Such practical and low-cost programs can bridge the widening violation-enforcement gap by helping authorities identify regular offenders and take action (which may include warnings or directed conversations, ticketing and fees).

The following paper sections describe (1) prior work on existing speed inference, vehicle license plate, color, make, and model recognition techniques; (2) our methods for estimating speeds and identifying vehicles; (3) application results in Austin, Texas; (4) a summary of survey findings on automated enforcement techniques in the U.S.; and (5) conclusions plus opportunities for future work.

## 2. SYNTHESIS OF RELATED WORK

Automated speed enforcement consists of two components: speed inference and vehicle identification. Speed inference serves as the basis for identifying potential speeders and provides estimates of speeds to determine if action needs to be taken. Once a vehicle is identified as speeding, vehicle identification is crucial for accurately and automatically identifying vehicles and extracting useful information for further use.

### 2.1 Speed Inference

Computer vision (CV) techniques to estimate vehicle speeds typically start by taking video recordings plus parameters (like image/size scaling factors) as inputs, then use detection and tracking algorithms to estimate distances traveled in a 2D (two-dimension/x-y) domain. Speeds are calculated with an estimated distance in the real world (calculated with a scale factor) and the time intervals between frames. Methods for distance calculation include photography-based (Kim et al., 2018), augmented intrusion line-based (Dahl and Javadi, 2019), pattern- or region-based, and image-based (Moazzam et al., 2019) techniques. Calibration plays a crucial role by helping calculate both intrinsic camera parameters (like sensor size, resolution, and focal length) and extrinsic parameters (such as location relative to the road surface). Vanishing points (VPs) (as shown in Figure 2, two VPs are accumulated separately by red and green edges) are commonly used for camera calibration and can be estimated using various algorithms, categorized into two main groups: geometry-based methods leverage the fact that VPs occur at the intersection of straight lines. These methods estimate VPs by associating lines to VPs (Feng et al., 2010), clustering lines (Barinova et al., 2010), or searching within a Gaussian sphere (Collins and Weiss, 1990). The second methods group focuses on learning to infer VPs from large-scale datasets containing VP annotations. For example, Zhai et al. (2016) extracted global image context with a deep convolutional network to constrain the location of possible VPs while Chang et al. (2018) trained models on one million Google street-view images to detect VPs. Based on estimated VPs and assumptions that the camera is free of skew and the principal point is at the center of the frame, the camera's intrinsic and extrinsic parameters can be calculated. These parameters enable a transformation between the camera's coordinate system and the world coordinate system. However, these methods are developed for fixed traffic cameras, which need further analysis in the case of mobile cameras.

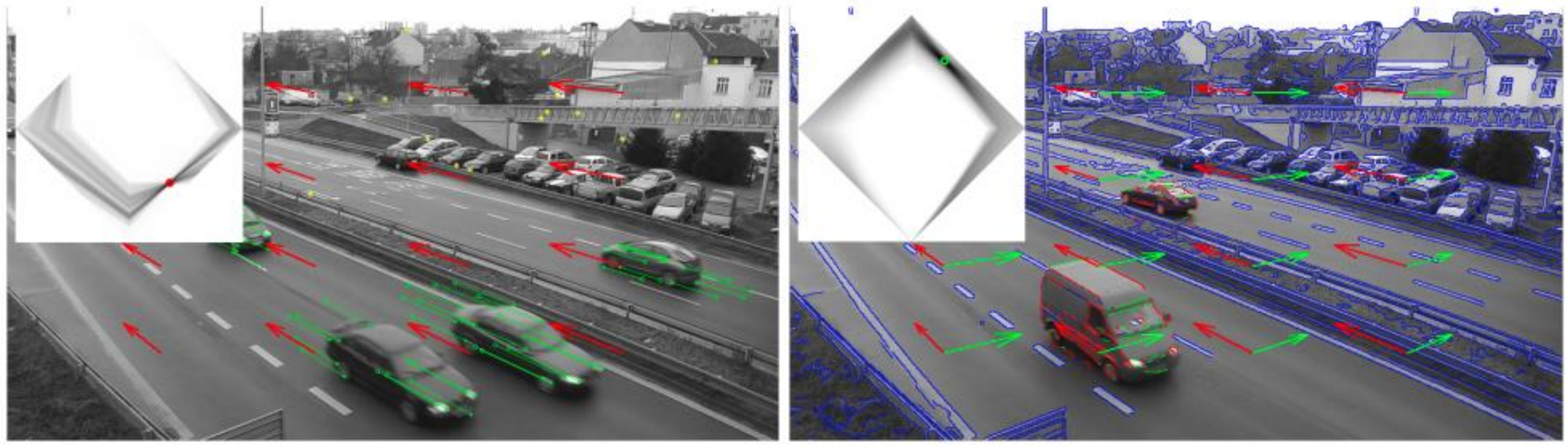

**Figure 2: Vanishing Points** (Source: Dubská et al., 2014)

### 2.2 Vehicle Identification

Vehicle detection algorithms are a type of object detection and are classified as one-stage detectors (such as You Only Look Once (YOLO) or Single Shot Detector (SSD)) or two-stage detectors (like Region with Convolutional Neural Network (R-CNN) and faster R-CNN). The latter use two neural networks to find and classify regions of interest, delivering greater accuracy but longer processing times (Kim et al., 2020). YOLO is a popular method for efficiently detecting vehicles and traffic violations, like jumping red-light signals (Ravish et al., 2021). Wang et al. (2023) analyzed the performance of YOLOv7 in detecting objects at different frame rates and found that it outperformed two-stage detectors in terms of both time and accuracy. Meanwhile, DeepSort (Wojke et al., 2017) is often used to track vehicles by adopting two association matrices (for object velocity and appearance) to create downstream-frame boxes via Kalman filters and then predicting vehicle positions across video frames.

License plates are essential for vehicle identification. After detecting and tracking vehicles, precise license plate identification can ensure that police tickets are delivered correctly. For instance, license plate recognition systems have been used for parking enforcement; they are installed on officer cars or at parking lot entrances and exits to scan and identify vehicles violating parking regulations. Automatic License Plate Recognition (ALPR) algorithms are the most common way to identify unique vehicles. This is a three-step process: first, the license plate is localized by either feature-based (Du et al., 2012) or deep learning-based (Laroca et al., 2019) methods. Then,character segmentation is done, and finally, recognition techniques are applied to extract the text. Current techniques use separate YOLO models to extract vehicles and license plates. Text recognition on these license plates is accomplished through either segmentation (a two-step process involving segmentation and a recognition model) or segmentation-free methods (a one-step process). There are several optical character recognition (OCR) techniques available (EasyOCR, 2021; Kuang et al., 2021; Pytesseract, 2022), which also pre-process images (de-skewing, smoothing edges, and converting images to black and white) to boost the chances of recognition (Karandish, 2019). ALPR algorithms are mainly hindered by poor image quality and low-resolution cameras. Much research has gone into improving image quality (Dong et al. 2015, Hamdi et al. 2021), and general adversarial networks (GANs) have proven successful in super-resolution reconstruction (Hamdi et al. 2021). While the entire pipeline used for ALPR on fixed camera videos (Silva and Jung, 2020; Zhang et al., 2021), including drone-recorded videos (Kaimkhani et al., 2022) is included in many publications, the accuracy and applicability of ALPR algorithms have not been validated for use with mobile phone video recordings.

License plate recognition may fail due to dark (nighttime or shade) conditions, occlusion by heavy rain or other vehicles, fake or missing plates, camera lens quality, and zoom level. In cases where a license plate is illegible, vehicle color, make, and model information can serve as alternative means to narrow down the possibilities of the vehicles involved in unlawful driving situations (Lee et al., 2019). Changing a vehicle's plate to commit crimes or avoid enforcement is relatively easy, but this is not the case for color, and especially not for make and model features. Proprietary tools are available for recognizing vehicle makes and models by using installed traffic cameras, but no such system exists for general phone cameras. Conversely, open-source approaches, especially application programming interfaces (APIs), are accessible and low-cost, helping institutions and communities worldwide reduce incidents of dangerous driving, death, and other losses. For example, PlateRecognizer (2024) advertises vehicle classification (including sedans, sports cars, pickup trucks, SUVs, etc.) across over 9,000 makes and models and is used in over 50 countries. RapidAPI (2024) detects vehicle color, make, model, generation, and orientation for more than 3,000 models common in the U.S. In terms of color detection, Baek et al. (2007) proposed an SVM (Support Vector Machines) method for color classification, and the implementation achieved a success rate of 94.92% for 500 outdoor vehicles with five colors (black, white, red, yellow, and blue). Tilakaratna et al. (2017) employed an SVM-based method with six features and provided a wide range of 13 colors for classification. Their method performs with an accuracy of 87.52% over 2,500 images.

## 3. VEHICLE SPEED DETECTION AND IDENTIFICATION

This paper assumes that mobile phones are held still while recording videos. Since videos are analyzed frame-by-frame, inclination angles and phone movements can be neglected over short time intervals.

### 3.1 Obtaining VPs and Estimating Speeds

In this work, VPs are obtained automatically in the first frame using Lu et al.'s (2017) detection algorithm. This algorithm iteratively and randomly selects two straight-line segments. It uses their intersection point as the first vanishing point ($V_1$) and then uniformly samples a second vanishing point ($V_2$) on the great circle or equivalent sphere of $V_1$, as shown in Figure 3. Starting from each VP, tangent lines of vehicle "blobs" (a group of pixels in a video frame representing a vehicle) are found, enabling the construction of 3D bounding boxes (Dubská et al., 2014). Using these two VPs, two lines are extended to intersect with the points inside the frame. Four intersection points from these extended lines provide a rectangle (with lines selected to avoid including at least one VP in the rectangle). Assuming the vehicles are moving toward one of the VPs, the perspective transformation can be constructed to rectify this rectangle, preserving only the vertical or horizontal movement of vehicles.

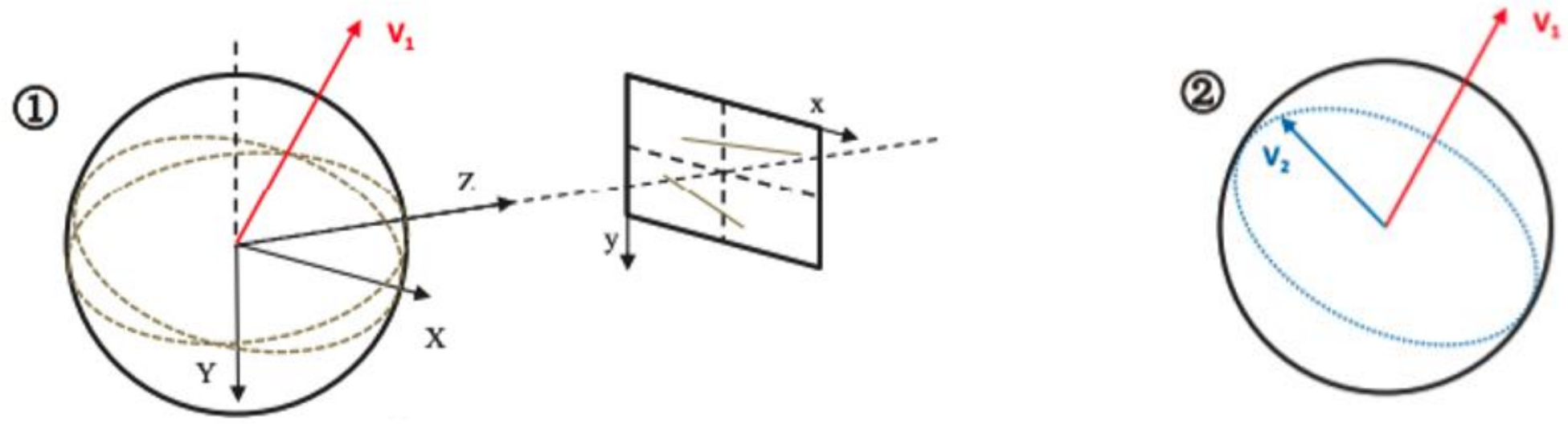

**Figure 3: Procedures of generating two VPs** (Source: Lu et al., 2017)

Denote two points at both ends of the 3D bounding box as $A=[a_x,a_y]^T$ ($a_x$, $a_y$ represent the x and y coordinates of point $A$ in the image plane) and $B=[b_x,b_y]^T$ in the former frame, and $A'=[a_x',a_y']^T$ and $B'=[b_x',b_y']^T$ in the next frame. Taking vehicle length ($L$) as a reference (assumed here as the median length of U.S. passenger vehicles, 4.5 meters (Ibiknle, 2024)), the actual moving distance ($x$) would be $||A-A'||\cdot L\,/\,||A-B||$. Vehicle speed estimate is then that distance ($x$) multiplied by frame rate, which is 30 frames/second (fps) for most smartphones.

### 3.2 Training Data for Vehicle Make and Model

Several datasets have been used to train models for automated make and model detection. For example, Yang et al.'s (2015) CompCar dataset consists of 136,727 internet vehicle images plus 44,481 surveillance-camera vehicle images across 153 car makes and 1,716 car models. Tafazzoli et al.'s (2017) Vehicle Make, Model Recognition Dataset (VMMRDb) was compiled across websites and contains 291,752 images for 9,170 distinct vehicle classes, but it ended with the 2016 model year. The average life span of U.S. passenger vehicles is roughly 16 years (Parekh and Campau, 2022), and this paper first identified the nation's 100 most popular vehicles from the 2017 National Household Travel Survey's (NHTS's) 220,430 million trip records (based on total vehicle-miles traveled by make/model). We scraped the Internet for 15,639 make/model images to use as a training dataset (alongside 300 images of those 100 most-used passenger-vehicle fronts, sides, and backs), as shown in Table 1.

**Table 1: Vehicle Make and Model Training Data**

| **Dataset** | **Training Data** |
|---|---|
| **# of Images in total** | 15,639 web-scraped images + 300 manually-collected images |
| **# of Images for each vehicle make/model** | 100-200 images per make/model (including front, back, and side views in different colors and settings) |
| **Method** | Collected automatically via web scraping & combed manually to remove irrelevant images. |
| **Example images** | 7 of 174 images for Ford F-Series |

### 3.3 Overall System Implementation

This paper relies on a series of deep-learning programs (as shown in Figure 4) for speed estimation and vehicle identification, incorporating object detection, object tracking, license plate recognition, make, model, and color detection to infer information from videos recorded via a mobile device. Vehicle bodies are first detected in each video frame using YOLOv8 code, then tracked/connected (across frames) using DeepSort and StrongSort (Du et al., 2023). Speeds are estimated via vehicle bounding boxes and VPs (as described above, in Section 3.1). Each cropped image of the tracked vehicles is sent to a fine-tuned YOLO v7 model for license plate detection (Anpr-Org, 2023). The detected and extracted license plate images are passed to a Super-Resolution Model (by Wang et al., 2018) and an Easy-OCR model (optical character recognition)(EasyOCR, 2021) to infer and output license plate characters. The ColorDetect (2024) package and histogram and Özlü's (2018) histogram and K-nearest neighbors (KNN) techniques are then used for color inference. The KNN method compares the bounding box image to 8 base colors (white, black, red, green, blue, orange, yellow and violet) and outputs the closest color match. Meanwhile, ColorDetect compares it to all possible RGB colors and provides the fraction of color present in the vehicle bounding box.

Meanwhile, the cropped image is sent to a Resnet-50 architecture model for make/model inference (He et al., 2016). This Convolutional Neural Network (CNN) model computes the dot product between two matrices, one representing features of images and another representing the convolutional 'kernel.' This process is accomplished by multiplying the corresponding values and adding the results to get a single scalar value in parallel (Taye, 2023), which helps preserve the spatial structures of images. It is large enough to capture variations in vehicle makes and models while also lowering computing time. In this work, the model is initially pre-trained on the VMMRDb dataset (Tafazzoli et al., 2017). Following pre-training, the last layer of the model is replaced with a fully connected layer with 100 nodes, to detect 100 top U.S. vehicle makes and models. The training dataset collected in Section 3.2 is then used to fine-tune and re-train the last layer. Freezing the earlier layers helps the model retain its learning from the VMMRDb dataset and the amount of data used for fine-tuning is relatively small compared to the amount CNNs usually need, so only the last layer is re-trained. In addition to using the VMMRDb dataset to pre-train the model, data augmentation is employed during fine-tuning to help increase the amount of data the model detects. Four transformations are used in this process; see Table 2. These transformations also deal with real-world issues like tilted videos, blurry recordings, dark environments, bad weather conditions, etc.

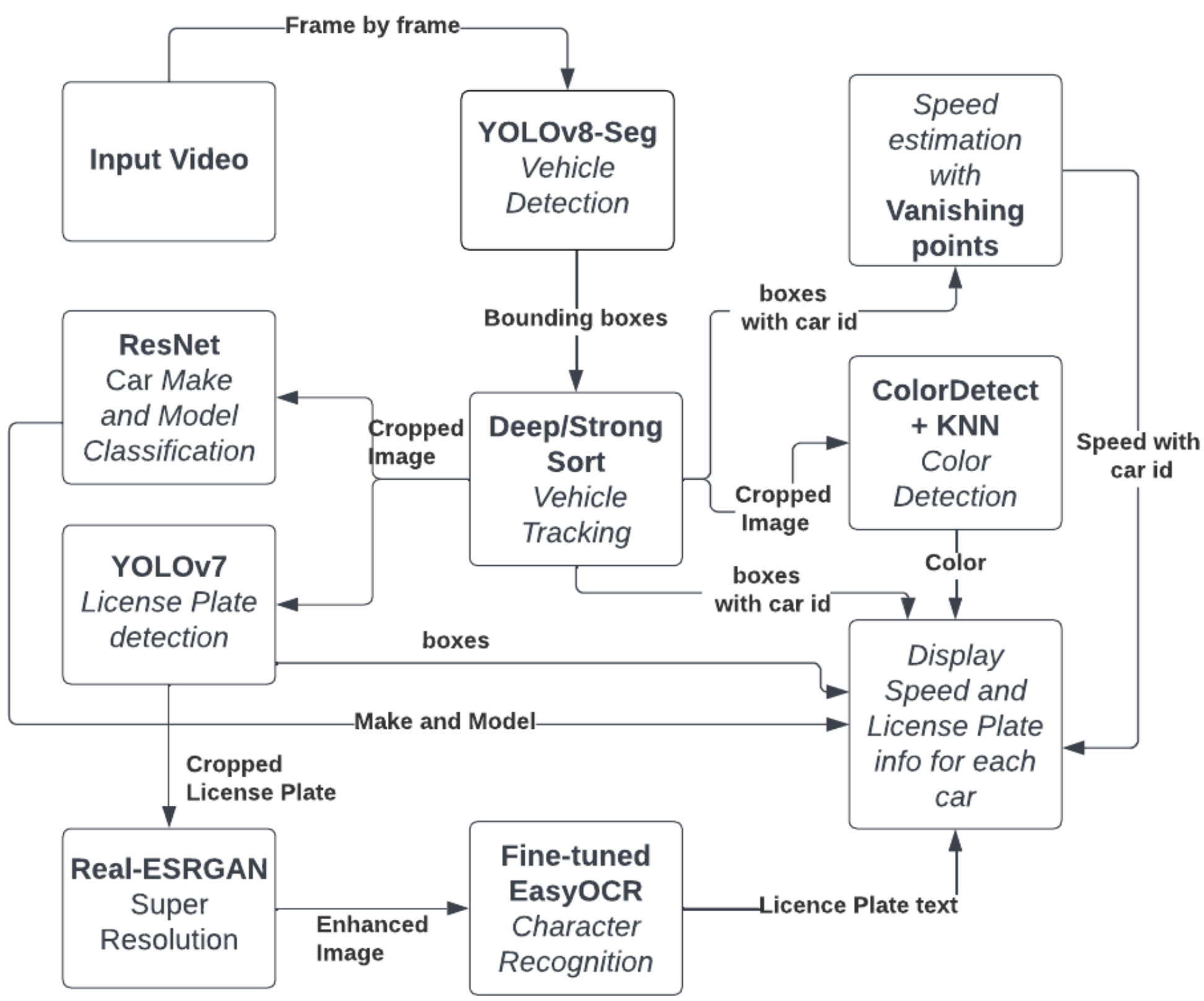

**Figure 4: Flowchart of Vehicle Speed Estimation and Identification System**

**Table 2: Four Transformations used for Data Augmentation** (Source: Butt et al., 2021)

| | Transformation | Purpose |
|---|---|---|
| | Gaussian Blur | Provide the model with different levels of blurred images. |

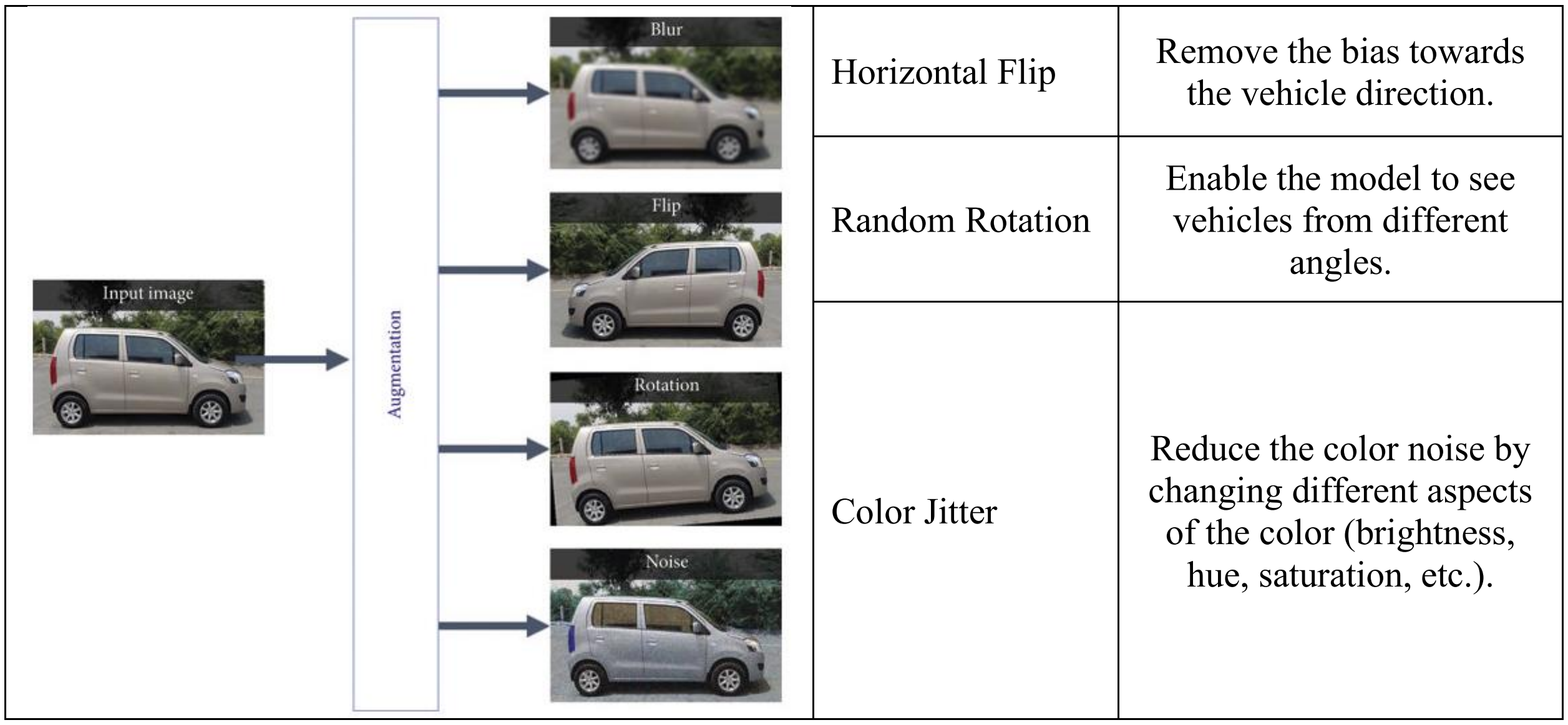

| | | |
|---|---|---|
| | Horizontal Flip | Remove the bias towards the vehicle direction. |
| | Random Rotation | Enable the model to see vehicles from different angles. |
| | Color Jitter | Reduce the color noise by changing different aspects of the color (brightness, hue, saturation, etc.). |

The crude results for each frame in the video include multiple features such as frame ID, vehicle ID, vehicle class, bounding box, color, make, model, license plate bounding box, and license plate text, as well as their respective probabilities. To streamline the analysis, a Python script was developed to process these results and generate a consolidated output for the entire video. The processed output includes a timestamp indicating when predictions are generated, vehicle ID for tracking all vehicles in the video, the most frequent vehicle class for each vehicle ID along with the mean probability, the mean speed, the most frequent color, the top 3 prevalent colors and their portions occupied in the image, the top 3 frequent makes and models with their mean probabilities, and the top 10 frequent license plates. A sample output is displayed in Table 3.

**Table 3. Sample Output of Vehicle Speed Estimation and Identification System**

| Output Image | Feature | Output |
|---|---|---|
| | Vehicle ID | 1 (first in frame[1]) |
| | Vehicle Class | Car |
| | Vehicle Class Probability (%) | 91% |
| | Speed (mi/hr) | 88.36 mi/hr |
| | Most Frequent Color[2] | Black |
| | Top 3 Prevalent Colors + Shares[3] | DarkSlateGray: 30.3%<br>Black: 25.6%<br>Gray: 16.5% |
| | Top 3 Makes & Models + Confidence | Mazda Rx8: 0.57<br>Honda Accord Sedan: 0.42<br>Ford Fusion: 0.38 |
| | Top 10 Plate Estimates + Confidence[4] | SLL[5]: 1.0, TOA: 0.96, SLL##35[6]: 0.95, SII##35: 0.86, TCAW: 1.0 |

Note:[1] One frame may include multiple vehicles. [2]Most frequent color is among 8 base colors in the KNN model. [3]Top 3 prevalent colors is among all possible RGB colors in the ColorDetect model. [4]When the number of predicted plates

is fewer than 10, the model will output all estimates. [5]Incomplete license plate estimates (like "SLL") are original predictions. [6]Hashtags (#) are to obscure actual values for photo anonymity. And SLL##35 is the correct prediction.

# 4. RESULTS

## 4.1 Performance of License Plate Recognition Model

The ANPR model was first tested on 1,800 images from the UFPR-ALPR dataset, a publicly available and commonly-used set of over 30,000 license plate characters from 150 vehicles (each with 30 images, 4500 images in total) captured in real-world scenarios in Brazil with a 30 FPS frame rate, where both the camera (the cameras used are: GoPro Hero4 Silver, Huawei P9 Lite and iPhone 7 Plus) and the vehicle are moving. The cameras are installed in another vehicle. Table 4 presents performance metrics for the YOLOv7 detection model in combination with either the Easy-OCR, Super Resolution (Real-ESRGAN) + Easy-OCR, or Super Resolution (Real-ESRGAN) + Fine-tuned Easy-OCR text recognition model. Figure 5 illustrates improvement of the Super Resolution technique. The Easy-OCR model's output is simply 'EE' - with no characters identified correctly, while Super Resolution predicts 'IU B6t5O62' - with 4 out of 7 characters identified correctly. Key reasons for low accuracy of the license plate text recognition model are the lack of clarity of extracted images and the fact that the Easy-OCR model is not specifically trained to recognize license plate characters. To further increase accuracy, the Easy-OCR model is fine-tuned on a small subset of UFPR license plates and synthetic data, reaching up to a 47.06% accuracy rate.

**Table 4: Performance of License Plate Recognition Model**

| Model | | Model Output | Criteria | # Correct Images | Accuracy |
|---|---|---|---|---|---|
| License Plate Detection | YOLOv7 | License plate bounding box. | The predicted bounding box covers more than 70% area of the true one. | 1413/1800 | 78.50% |
| License Plate Text Recognition | Easy-OCR | License plate characters. | The predicted license plate is the same as the true one. | 252/1800 | 14.00% |
| | Super Resolution (Real-ESRGAN) + Easy-OCR | | | 407/1800 | 22.61% |
| | Super Resolution (Real-ESRGAN) + Fine-tuned Easy-OCR | | | 847/1800 | 47.06% |

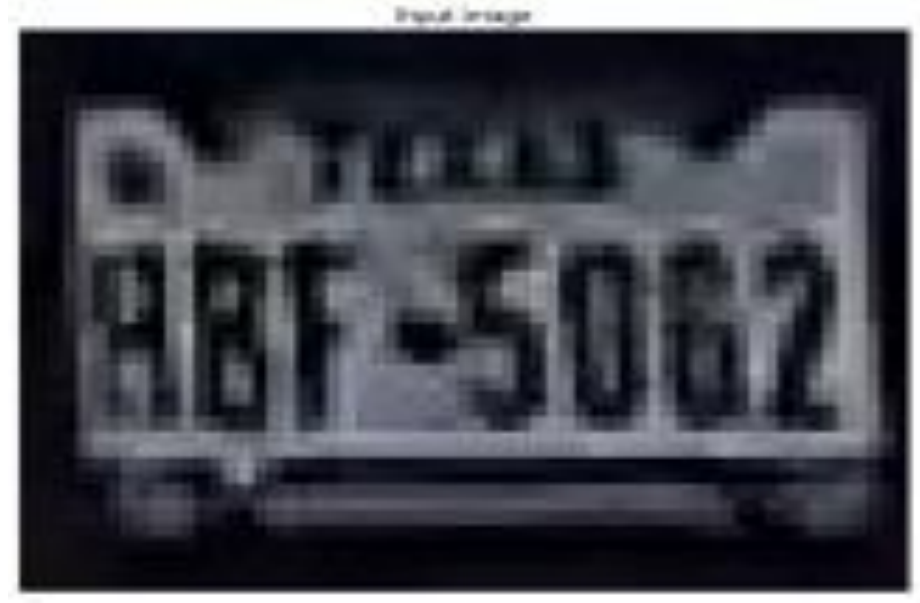

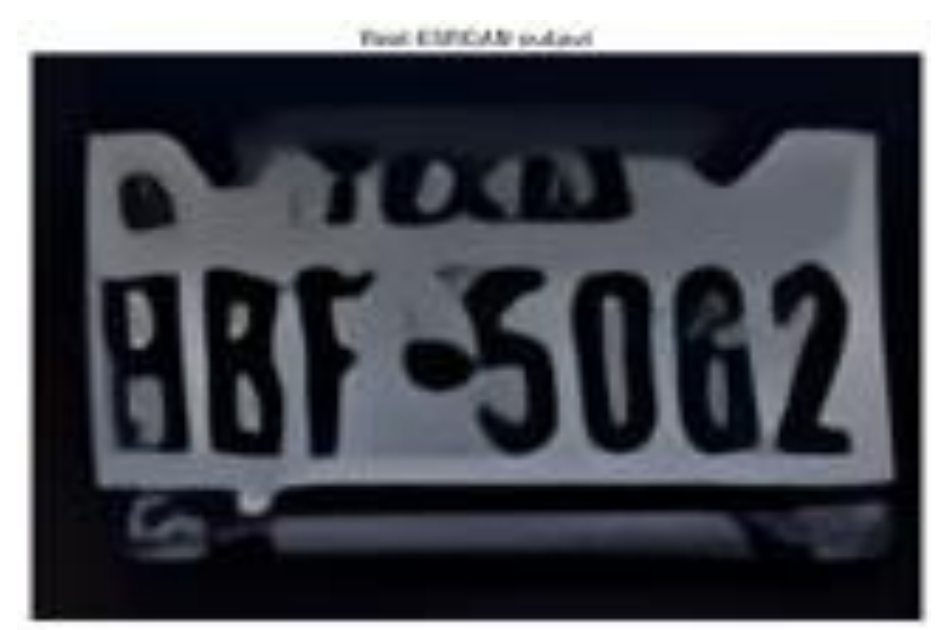

(a) Output of Easy-OCR (b) Output of Super Resolution + Easy-OCR

**Figure 5: ANPR Improvement using Super Resolution**

### 4.2 Performance of Overall System

To assess the overall system's performance in estimating speeds and identifying plate, make, model, and color, this work collected 73 smartphone-recorded videos (4 to 5 second durations each, within 1.5 miles of the University of Texas at Austin campus) and 42 traffic-camera recordings (2 to 3 seconds each, at Austin intersections). These 115 videos contained reasonable imagery of 148 separate vehicles during the daytime, and accurate make, model, color, and plate information could be obtained by eye (human/manual review of the videos) or from images of slowed vehicles downstream at a red signal light. "True" speeds for these 148 vehicles were determined using speed radar guns or image-frame-by-frame review. Manual frame review was also used to provide make, model, color, and plate numbers. Table 5 displays accuracies for each feature, with color identification around 60.8% accuracy. The combined color codes from the KNN model and the Detect model excelled in distinguishing gray and black vehicles. However, they tended to confuse other paint and body colors because the codes detect the colors of the entire bounding box, which includes tires, rims, and other parts of the car body. Vehicle manufacturer (model) identification was 48.6% accurate, and model was just 16.9% accurate when using the Top 3 model estimates. Speed (within 20% of "true" speed) and license plate (excluding the state character) were 16.3% and 29.7% accurate, respectively, due to factors like parked cars and handheld/moving or blurry phone-camera images.

**Table 5. Performance Results of Overall System**

| Features | Criteria | #Correct/ Sample Size | % Correct |
|---|---|---|---|
| Color | Predicted color(s) is correct. (Out of all possible RGB colors) | 90/148 | 60.81% |
| Make | Actual make is in Top 3 predictions. | 72/148 | 48.65% |
| Make & Model | Actual make + model occur among Top 3 predictions. | 25/148 | 16.89% |
| Speed | Predicted speed within 20% of actual. | 24/147 | 16.33% |
| License Plate | Actual license plate is among Top 10 predictions. | 30/101[1] | 29.70% |

Note: [1]License plates are unreadable in 47 testing vehicles.

## 5. SURVEY FINDINGS

To supplement these numeric results, an online survey was distributed to US law enforcement agency officers across US states. The survey asked for participants' thoughts regarding 1) major challenges for automated enforcement application inside the US, 2) best applications they have seen for automated enforcement (anywhere in the world), 3) use of individuals' smartphones to assist US law enforcement practices, and 4) automated enforcement accompanied by automated ticketing of other/non-speeding behaviors (like illegal parking).

Their responses highlight the effectiveness of automated enforcement systems (in the U.S. and elsewhere), with Europe's time-over-distance (average speed) camera systems and the U.S.'s speed + red-light cameras proving effective and defensible. Table 6 shows responses relating to top challenges, with public perception, privacy, safety, and practicality listed as top concerns.

**Table 6. Concerns + Challenges in U.S.-based Automated Enforcement**

| Area | Challenges |
|---|---|
| Public Perception | • Public may be unaware of automated enforcement's benefits.<br>• Automated enforcement got off on the wrong foot in the US and looked too much like a money grab by local governments and the automated industry. It needs to be revenue neutral, focused on safety, with industry kept on a short leash. |
| Privacy and Related Topics | • Privacy concerns. The vehicle information may be revealed for some commercial use.<br>• Emergence of public vigilantes. There may be possible abuse in submitting videos, like swapping fake plates via AI methods. |
| Practicality in Application | • Location of cameras is challenged; law enforcement agencies need to involve communities in site selection and support the locations by being transparent with data.<br>• Officers conducting speed enforcement will eventually have to testify to their training and calibration of equipment used.<br>• Emergency vehicles should be exempt.<br>• Driver Distractions. In most driving situations, speed naturally increases downhill and decreases uphill; decreases in congested traffic, increases in the absence of traffic, and so on. If the driver is paying too much attention to the speedometer, he may be failing to pay attention to the road ahead, causing more crashes than are prevented with speed enforcement technology.<br>• Possible vulnerability that the system may be filled with unnecessary submissions. |

Using computer vision with smartphone images to assist in making roadways safer, via follow-up enforcement, appears both promising and natural. It is much like anyone calling 911 or another emergency hotline to report what they see with their own eyes, but provides the advantage of safety officers being able to review the footage themselves. Several participants recommended working to obtain public buy-in, and making such video submissions part of a larger safety campaign, where the objective is not revenue but safety. For example, privately-provided images may simply be used to increase police patrol of certain locations at certain times of the week or year, as should be done when individuals leave messages with 911 and 311 operators in the US every day.

Compared to installing cameras for automated enforcement, built-in speed governors appear to be the easiest way to limit vehicle speeds. As of now, most U.S. fleet owners install speed governors on heavy-duty trucks to ensure safety. Although the cost of built-in speed governors is not well-calculated, a standard Automated Emergency Braking with Forward Collision Warnings, Lane Departure Warnings, and Adaptive Cruise Control system is estimated to cost a fleet over $4,000 (FMCSA, 2024). Estimating the costs for medium-duty trucks is more complex due to necessary vehicle modifications and older chassis that may lack the wiring in place for some sensors or driver interfaces. Speed governors on trucks are implemented through electronic control units (ECU), which can be set at the factory, or changed by OEM-specific software. Nevertheless, one hidden challenge arises when the vehicle needs to deal with significant segment-based speed limit changes. In such scenarios, imperfect functioning of segment-based speed limiters could place drivers in difficult situations.

While large fleets are more likely to use speed governors, most fleets in the U.S. are small or independent. Some drivers and car manufacturers may be unhappy with any system that might reduce vehicle performance. In this context, a monitoring-only application would be more suitable. For instance, the Life360 app offers paid location-based services that can generate reports for monitoring driving behaviors. Smartphones, equipped with necessary sensors (such as GPS and accelerometers), can measure vehicle speeds and accelerations. Therefore, it is technically feasible to develop an app that tracks and reports speed versus set speed limits. However, this issue is more about market interests and politics. Additionally, maintaining such a system would require a comprehensive database of road segments and speed limits.

## 6. CONCLUSIONS AND FUTURE WORK

This study demonstrates the potential and practicality of a smartphone-based method in the context of automated speed enforcement to improve road safety. When tested on 1,800 images from Brazil's UFPR-ALPR dataset, the license plate number recognition model detected 78.5% of license plates and accurately recognized 47.1% of license plates' text. The entire system achieves 16.33% accuracy in estimating speeds, with errors staying within a 20% range, and 29.70% in recognizing license plates. As a supplement to identifying vehicles, it can reach up to 60.81% and 48.65% accuracy when detecting vehicle colors and makes, respectively. This research aims to envision further individual engagement in regulating traffic laws and involvement of autonomous technologies in this process. It is evident that these technologies can play a pivotal role in enhancing road safety and traffic management, and additional research will be key to realizing these goals. This work also investigates the common challenges of automated enforcement and future hurdles and recommendations of the practical use of private recordings for enforcement purposes.

To improve speed estimation accuracy, the specific length of each vehicle (by make/model) should be used instead of a single average or median assumption, as is currently used (especially for very long or unusually short vehicles). The model for color detection can be modified to focus on specific parts of the vehicle, such as the hood and trunk, rather than considering the entire image within the bounding box. The entire system can be made more accurate by training the model with moving camera data, collecting and labeling more data, and working to eliminate noise from nearby vehicles. Identification of a vehicle's make, model, year, and color will prove useful when license plates are obstructed or missing (or falsified), increasing the likelihood of successful law enforcement for safer roadways. Mobile camera properties, like aperture size and shutter speed, can be experimented with to improve video recordings without motion blur.

Directions for future research include extending the analysis to more complex scenarios such as nighttime videos (in lighted and unlighted settings) when speed and plate inference will probably prove more difficult. Additionally, research should focus on scenarios with moving cameras, as is common with hand-held devices or when inside nearby vehicles. Another extension is developing a mobile smartphone application for regular or automated submission of flagged video segments with precise position/location details (during actual recording rather than from user-estimated values). The scalability of the presented idea needs to be explored to see how it will perform in dense observation environments, such as expressways and city centers.

Another endeavor is building comprehensive maps for relevant enforcement agency response. Encouraging enforcement agencies to adopt private-phone video for enforcement support may be challenging due to data privacy concerns, as well as the potential for fake video submissions. However, automakers like General Motors are already surveilling and sharing such driving behavior with insurance companies. Currently, many US states do not allow the use of traffic cameras or speed cameras for law enforcement purposes, but other nations rely heavily on and benefit greatly (in terms of safety, effort, and cost) from automated enforcement. From a system implementation standpoint, an end-to-end system that can optimize the current system is preferred. Automating the entire system decreases human involvement and manual costs.

## ACKNOWLEDGEMENTS

We are grateful to UT Austin's Good Systems research program for sponsorship, Eva Natinsky and Samridhi Roshan for helping generate license plate results, and undergraduates Connor Moening and Austin Turgeon for trimming and labeling smartphone videos. We thank undergraduate Eric McDonnell for his early review of state laws, Tong Wang and Xuxi Chen for help with the speed estimation model, Aditi Bhaskar, Lucas Allison, Ali Khanpour, and Robert Lehr for their valuable edits, and Sheila Lalwani and Dr. Mike Murphy for information on courtroom admissibility of phone-recorded videos.